\documentclass[conference,10pt]{IEEEtran}
\IEEEoverridecommandlockouts
\usepackage{cite}
\usepackage{amsmath,amssymb,amsfonts}
\usepackage{algorithmic}
\usepackage{graphicx}
\usepackage{textcomp}
\usepackage{xcolor}
\usepackage{booktabs}
\usepackage{multirow}
\usepackage{subcaption}    

\begin{document}


\title{CLAD: A Clustered Label-Agnostic Federated Learning Framework for Joint Anomaly Detection and Attack Classification}

\author{
\IEEEauthorblockN{
Iason Ofeidis\IEEEauthorrefmark{1},
Nikos Papadis\IEEEauthorrefmark{2},
Randeep Bhatia\IEEEauthorrefmark{2},
Leandros Tassiulas\IEEEauthorrefmark{1},
TV Lakshman\IEEEauthorrefmark{2}
}
\IEEEauthorblockA{\IEEEauthorrefmark{1}
Yale University, New Haven, CT, USA
}
\IEEEauthorblockA{\IEEEauthorrefmark{2}
Nokia Bell Labs, Murray Hill, NJ, USA
}

}

\maketitle

\begin{abstract}

The rapid expansion of the Internet of Things (IoT) and Industrial IoT (IIoT) has created a massive, heterogeneous attack surface that challenges traditional network security mechanisms. While Federated Learning (FL) offers a privacy-preserving alternative to centralized Intrusion Detection Systems (IDS), standard approaches struggle to generalize across diverse device behaviors and typically fail to utilize the vast amounts of unlabeled data present in realistic edge environments. To bridge these gaps, we propose CLAD, a holistic framework that seamlessly incorporates Clustered Federated Learning (CFL) with a novel Dual-Mode Micro-Architecture ($\text{DM}^2\text{A}$). This unified approach simultaneously tackles the two primary bottlenecks of IoT security: device heterogeneity and label scarcity. The $\text{DM}^2\text{A}$ component features a shared encoder followed by two branches, enabling joint unsupervised anomaly detection and supervised attack classification; this allows the framework to harvest intelligence from both labeled and unlabeled clients. Concurrently, the clustering component dynamically groups devices with congruent traffic patterns, preventing global model divergence. By carefully combining these elements, CLAD ensures that no data is discarded and distinct operational patterns are preserved. Extensive evaluations demonstrate that this integrated approach significantly outperforms state-of-the-art baselines, achieving a 30\% relative improvement in detection performance in scenarios with 80\% unlabeled clients, with only half the communication cost.

\end{abstract}

\begin{IEEEkeywords}
Federated Learning, Internet of Things, Clustered Federated Learning, Multi-task Learning, Semi-supervised Learning, Intrusion Detection, Heterogeneity, Anomaly Detection, Attack Classification, Personalization
\end{IEEEkeywords}

\section{Introduction}









The integration of the IoT and IIoT has embedded intelligent connectivity into the fabric of critical infrastructure, smart cities, and autonomous systems~\cite{al2015internet, cisco2020cisco, sisinni2018industrial}. As these networks expand, they create a vast and diverse attack surface, ranging from low-power environmental sensors to high-performance autonomous drones~\cite{antonakakis2017understanding}. Securing these ecosystems is paramount; however, the sheer scale and heterogeneity of IoT devices make traditional defense mechanisms increasingly obsolete~\cite{neshenko2019demystifying}. 

Traditionally, network security has relied on centralized Intrusion Detection Systems (IDS), where data from all edge devices is transmitted to a central server for analysis~\cite{liao2013intrusion}. In the IoT context, this approach can be fundamentally flawed. Centralization creates significant bottlenecks in scalability and bandwidth~\cite{bonawitz2019towards}, introduces substantial latency for real-time applications, and, perhaps most critically, violates strict privacy standards by exposing sensitive operational data to external servers~\cite{mothukuri2021survey}.

\begin{figure}[t]
    \centering
    \includegraphics[width=\linewidth]{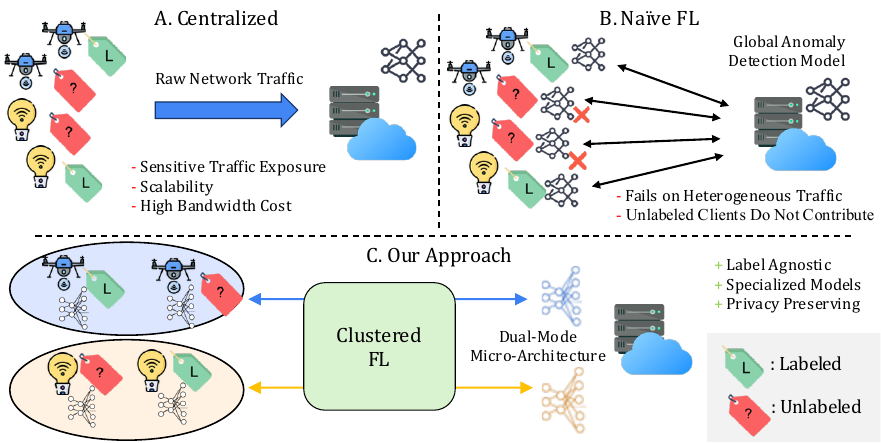}
    \caption{While centralized methods compromise privacy and standard FL fails to utilize unlabeled heterogeneous traffic, our proposed framework employs clustering and $\text{DM}^2\text{A}$ to ensure specialized, label-agnostic network anomaly detection and attack classification.}
    \label{fig:introduction}
\end{figure}

Federated Learning (FL) has emerged as a solution to these centralized limitations, enabling devices to train collaborative models by sharing only weight updates rather than raw data~\cite{mcmahan2017communication}. While FL addresses the privacy and bandwidth issues, standard implementations underperform in IoT environments because they force a single global model to represent a highly heterogeneous population of devices~\cite{zhao2018federated, li2020federated}. A traffic pattern that is benign for a high-throughput video camera may mimic a Denial-of-Service (DoS) attack for a constrained sensor. As illustrated in Figure~\ref{fig:introduction}, a na\"ive global FL model struggles in distinguishing these contradictory patterns, often resulting in suboptimal model performance.

To address this, Clustered Federated Learning (CFL) serves as a critical architectural evolution. CFL segments the network, dynamically grouping devices with similar behaviors to train specialized, high-accuracy models~\cite{sattler2020clustered, ghosh2020efficient, fan2024taking}. Unlike standard FL, which blindly aggregates updates from incongruent sources, CFL identifies communities of clients with congruent data distributions (e.g., separating ``camera traffic'' from ``sensor traffic''). This separation is indispensable in complex IoT ecosystems, as it ensures that the unique operational patterns of specific device groups are preserved rather than averaged out. By maintaining distinct models for distinct clusters, CFL resolves the conflicts inherent in global aggregation, leading to faster convergence and drastically improved performance.

However, a distinct and often overlooked limitation in existing literature is the rigid adherence to single-task learning. Most FL frameworks (standard or clustered) target either supervised attack classification or unsupervised anomaly detection in isolation. This ``one-or-the-other'' approach is insufficient because it fails to account for the inherent differences in data availability across clients. In realistic IoT environments, obtaining high-quality labels is expensive and rare, leading to a scenario where a few clients possess labeled data while the majority operate with unlabeled traffic~\cite{10004993, yang2022survey}. Consequently, in existing supervised systems, unlabeled clients do not contribute to learning performance improvement, while purely unsupervised systems are blind to the specific attack intelligence offered by labeled ones. To achieve operational resilience, an IDS must be capable of utilizing all information available from participating devices. It must seamlessly integrate supervised and unsupervised learning streams, ensuring that no label, however scarce, is discarded, and no device, however unlabeled, is excluded from the collective defense.

To bridge these gaps, we propose CLAD, a \underline{C}lustered \underline{L}abel-agnostic federated learning framework for joint \underline{A}nomaly \underline{D}etection and attack classification that unifies Clustered FL with a Dual-Mode Micro-Architecture ($DM^2A$). Our lightweight $DM^2A$ model features a shared encoder that branches into two tasks: unsupervised anomaly detection and supervised attack classification. This design allows us to decouple the dependency on labels. Clients with labeled data refine the network's classification capabilities, while clients with only unlabeled data contribute to the underlying anomaly detection boundaries. This hybrid approach extends security coverage across the network. By enabling exchange of knowledge within clusters, a simple, unlabeled sensor can effectively benefit from the threat signatures learned by a sophisticated, labeled peer with a similar behavioral pattern. Critically, the framework is engineered for efficiency and robustness. It minimizes communication and computational overheads, ensuring viability on resource-constrained devices. The key contributions of this paper are summarized as follows:

\begin{itemize}
    \item \textbf{A Unified Dual-Mode Framework}: We propose CLAD, a lightweight joint learning framework capable of simultaneous Anomaly Detection and Attack Classification. The unique Dual-Mode Micro-Architecture allows resource-constrained IoT devices to perform both tasks efficiently within a single training pass.
    
    \item \textbf{Hybrid Label Utilization}: We introduce and validate a strategy to leverage both labeled and unlabeled data, maximizing data utility and performance in label-scarce environments.
    
    \item \textbf{Personalization via Clustering}: We demonstrate that grouping similar clients prevents global model confusion, significantly boosting learning performance.
    
    \item \textbf{Robustness and Efficiency}: Our approach outperforms state-of-the-art baselines, while exhibiting high accuracy and low communication and computational overhead and maintaining high performance even with small local datasets and varying network scales.
    
    \item \textbf{Algorithm-Agnostic Flexibility}: 
    Our framework is modular and supports the integration of various CFL algorithms to adapt to different IoT environments.
\end{itemize}

The rest of the paper is structured as follows. Section~\ref{sec:related_work} reviews related work. Section~\ref{sec:system} introduces our proposed model and framework. Section~\ref{sec:experimental_setup} outlines the experimental setup and Section~\ref{sec:results} presents the results. Finally, Section~\ref{sec:conclusion} concludes the paper.

\section{Related Work}
\label{sec:related_work}


\textbf{Centralized intrusion detection}
Most IDSs are designed under the strong assumption that data resides in a centralized repository.
For instance, \cite{rabbani_device_2025} simultaneously performs IoT device identification and anomaly detection using a feature extraction process and a centralized classifier.
In a similar centralized fashion, \cite{bhatia_unsupervised_2019} introduces an unsupervised learning approach based on autoencoders and Principal Component Analysis (PCA) to identify both known and unknown anomalies.

\textbf{FL for Anomaly Detection}
The assumption that the data can be centralized is often not realistic, as data/device owners might be reluctant to share raw data.
D{\"I}oT \cite{nguyen2019diot} is probably the first work that applied FL for anomaly detection. 
It relies on an external method for device type identification, and its goal is device-type-specific anomaly detection with no false alarms.
This is achieved with a hierarchical architecture: model training and anomaly detection (inference) happens at ``security gateways'' (akin to edge servers), and federated averaging happens at the ``IoT security service'', i.e. centrally.
Unlike D{\"I}oT, our approach does not rely on known or precomputed device types.
Instead, it groups devices with similar behavior through clustering that happens intertwined with model training.
Mothukuri et al. \cite{mothukuri_federated_2022} performs regular FedAvg (i.e. the first FL algorithm as introduced by \cite{mcmahan2017communication}) on Gated Recurrent Units (GRUs) and utilizes ensemble learning to achieve attack classification in a supervised setting.
Jithish et al. \cite{jithish_distributed_2023} performs an experimental study to compare FedAvg's performance when using different models (regression, autoencoders, classifiers) for anomaly detection in IoT and smart grids.
Kelli et al. \cite{kelli_ids_2021} creates an FL-based IDS that utilizes active learning to achieve personalization.

\textbf{FL for Attack Classification }For certain application domains, detecting anomalies may not be enough, since different attacks require different mitigation approaches.
Thus, attack type identification becomes necessary.
Some works have looked at this problem from a supervised lens.
For instance, LocKEdge \cite{huong_lockedge_2021} applies FL using a simple neural net to classify labeled DoS and other attack data. FedJam~\cite{panitsas2025fedjam} similarly leverages a multimodal lightweight FL framework to classify wireless jamming attacks, while \cite{zainudin_federated_2023} uses a more elaborate architecture to perform supervised attack classification for Software-Defined Networks (SDN)-based IIoT environments.  
A different approach by \cite{aouedi_federated_2023} utilizes FL in a semi-supervised setting: an autoencoder is trained using unlabeled data in a federated manner, and the server adds some layers to the model and trains it in a supervised and centralized manner on publicly available data.


\textbf{Clustered FL for Anomaly Detection}
Clustered FL has been employed by some works for anomaly detection purposes in an IoT setting.
ClusterFLADS \cite{fan2024taking} performs regular federated averaging as per \cite{mcmahan2017communication} for a few rounds.
It then uses feature extraction on the local models via PCA, and the extracted feature vectors are clustered using k-means.
From then on, multiple federated averaging algorithms are run in parallel, one per cluster.
Wei et al. \cite{wei_toward_2025} introduces a one-shot clustering method that groups clients into clusters, and uses a committee of clients in order to assign scores to models and filter out outlier models before averaging.
In \cite{saez2023clustered}, clients send their model parameters to the server, the server applies PCA on them and finds the best number of clusters by sweeping through a range of possible values. Then, each of the formed clusters performs FedAvg.
Unlike the above one-shot clustering approaches, our approach performs simultaneous clustering and training, i.e. the clustering is refined at every round, resulting in higher clustering and model accuracy.

Overall, our approach extends the existing literature by performing joint anomaly detection and attack classification thanks to our dual-mode architecture, and doing so in a personalized manner that is friendly to different clustered FL algorithms, while taking advantage of both labeled and unlabeled data.

\section{System Architecture}
\label{sec:system}

We now introduce CLAD's architectural components that allow it to leverage both labeled and unlabeled data to jointly perform the two tasks in a personalized fashion.
We consider an environment with $N$ devices belonging to $K$ different groups (e.g. of the same device type or behavioral pattern).
Each client $i \in \{1,\dots, N\}$ has a private dataset $\mathcal{D}_i$ consisting of potentially both benign and attack samples.
The system consists of two primary components: a \textit{Dual-Mode Micro-Architecture ($\text{DM}^2\text{A}$)} deployed on edge devices, and the employment of the \textit{CLoVE} mechanism \cite{bhatia2025clove} on the server for clustered federated aggregation\footnote{Note that we opt against using tree-based models (e.g. decision trees, random forests, gradient-boosted decision trees), as they face notable challenges in FL settings: significant communication overhead to determine optimal splits at each node and scaling difficulties as the number of clients/features increases \cite{Wang2024Decision, Lim2024A, Qian2025Tree-based}; complex aggregation as each client's tree structure may vary based on its local data \cite{Wang2024Decision, Lim2024A, Argente-Garrido2025An}; and sensitivity to data heterogeneity, resulting in uneven predictive performance across clients \cite{Lim2024A, Gao2024Balancing}.}.

\subsection{Dual-Mode Micro-Architecture ($DM^2A$)}
\label{sec:system_model}

To reconcile the competing requirements of high-fidelity anomaly detection and attack classification while achieving resource efficiency, we propose the \textbf{Dual-Mode Micro-Architecture ($\text{DM}^2\text{A}$)}, where \textit{micro} denotes a compact, lightweight design tailored for constrained IoT devices. This framework acts as a unified learning engine capable of toggling between unsupervised anomaly detection and supervised attack classification.

\begin{figure}[t]
    \centering
    \includegraphics[width=0.8\linewidth]{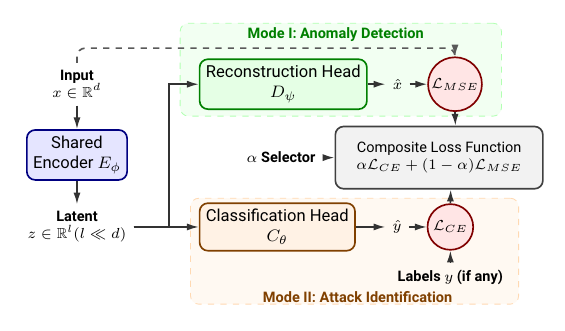}
    \caption{The proposed Dual-Mode Micro-Architecture ($DM^2A$). 
    }
    \label{fig:placeholder}
\end{figure}

\begin{figure*}[t]
    \centering
    \includegraphics[width=\linewidth]{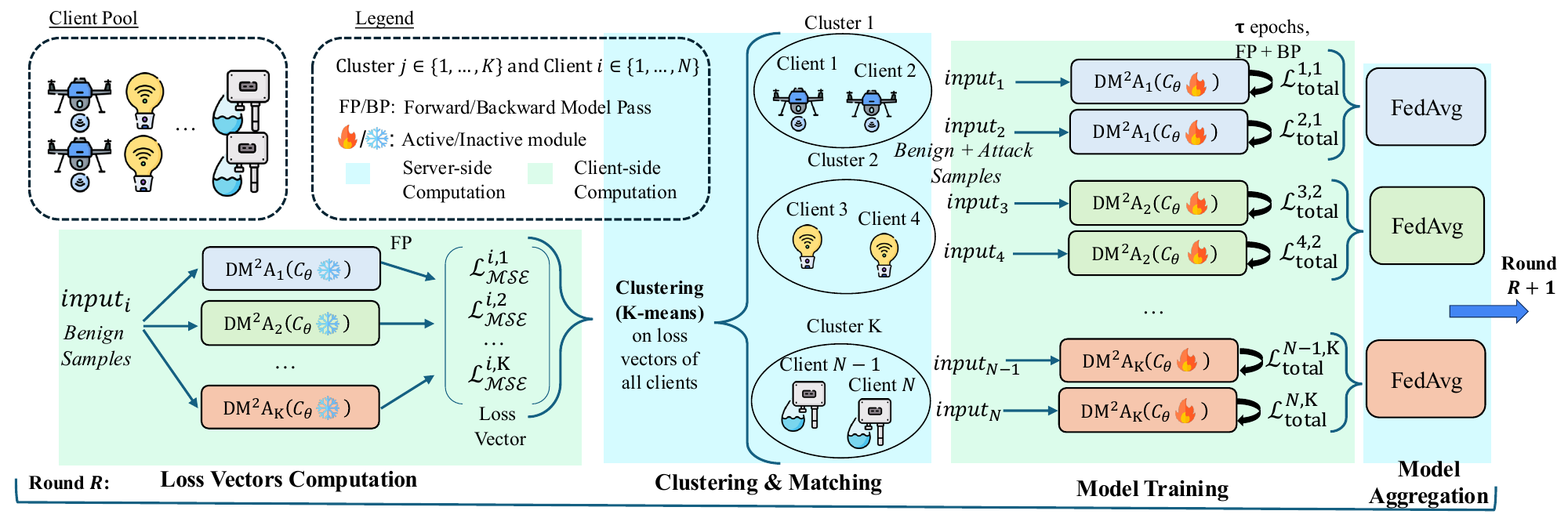}
    \caption{System Overview}
    \label{fig:system_overview}
\end{figure*}

The $\text{DM}^2\text{A}$ design consolidates feature extraction and task execution into three integrated sub-modules:

\noindent \textbf{Shared Encoder ($E_\phi$):} 
Maps high-dimensional traffic features $x \in \mathbb{R}^d$ to a compressed latent representation $z \in \mathbb{R}^l$ (where $l \ll d$) through the model weights $\phi$. By sharing this backbone across modes, $\text{DM}^2\text{A}$ eliminates redundant parameters and enforces a \textit{dual-constraint} on the latent space: $z$ must be simultaneously \textit{discriminative} (separating classes) and \textit{generative} (preserving structure). This is critical for our semi-supervised motivation: even without labels, the encoder captures meaningful traffic patterns via the reconstruction task, preventing the model from degrading when labeled data is scarce.

\noindent \textbf{Dual-Mode Operation:}
The latent vector $z$ feeds into two parallel branches. This design serves two strategic purposes: (a) \textit{Operational Flexibility}, allowing the device to adapt to label availability, and (b) \textit{Disentangled Loss Representation}, which enables us to utilize the reconstruction loss specifically for clustering logic (see Sec.~\ref{system:B}) rather than na\"ively relying on the composite loss.

\begin{itemize}
    \item \textbf{Mode I: Anomaly Detection (Reconstruction Head $D_\psi$):} A decoder mapping $z$ back to the input space through the weights $\psi$, producing $\hat{x} = D_\psi(z)$. This mode enables unsupervised learning on unlabeled data, establishing a baseline of ``normality'' for anomaly detection.
    \item \textbf{Mode II: Attack Identification (Classification Head $C_\theta$):} A dense layer mapping $z$ through the weights $\theta$ to a probability distribution over known classes, $\hat{y} = C_\theta(z)$. This mode fine-tunes decision boundaries for specific threats using available labeled samples.
\end{itemize}


\noindent \textbf{Composite Loss Function:}
For each client $i \in \{1, \dots, N\}$ and model $M_j$, $j \in \{1, \dots, K\}$, we minimize a composite objective $\mathcal{L}_{\text{total}}$ that balances the two operational modes:

\begin{equation}
\mathcal{L}_{\mathrm{total}} = \alpha \mathcal{L}_{\mathrm{CE}}(y, \hat{y})  + (1 - \alpha) \mathcal{L}_{\mathrm{MSE}}(x, \hat{x})
\end{equation}
We denote $\mathcal{L}_{\mathrm{total}}^{i,j} = \mathcal{L}_{\mathrm{total}}(\mathcal{D}_i, M_j)$,
where $\mathcal{L}_{\mathrm{CE}}$ is the Cross-Entropy loss and $\mathcal{L}_{\mathrm{MSE}}$ is the Mean-Squared Error loss, and $\alpha \in [0, 1]$ acts as the \textit{mode selector}.
This formulation provides architectural flexibility: in scenarios where labels are unavailable, $\alpha$ is set to $0$, reducing the task to pure anomaly detection via reconstruction error. When labels are present, each client can set their own $\alpha$ value based on their label availability to allow both heads to update the shared encoder accordingly.

\subsection{Federated Training with Adapted CLoVE}
\label{system:B}


Standard FedAvg fails in heterogeneous IoT environments because averaging gradients from functionally distinct devices results in negative transfer. Consequently, we select to employ the \textit{CLoVE} (Clustering of Loss Vector Embeddings) framework~\cite{bhatia2025clove}, due to its demonstrated efficacy over other CFL approaches. We adapt this framework to our anomaly detection and attack classification context with two domain-specific constraints:
\begin{enumerate}
    \item \textbf{Reconstruction-Only Fingerprinting:} We utilize only the Reconstruction head ($D_\psi$) and freeze the Classifier head ($C_\theta$) (indicated by the snowflake symbol in Figure~\ref{fig:system_overview}) during loss vector computation. This ensures clients are grouped based on traffic structure rather than label distribution.
    \item \textbf{Benign-Sample Filtering:} We compute loss vectors using only the subset of benign samples ($\mathcal{D}_i^{\mathrm{benign}} \subset \mathcal{D}_i$). This prevents clustering skews caused by transient attacks.
\end{enumerate}

The complete training protocol can be seen in Figure~\ref{fig:system_overview}  and proceeds as follows:

\noindent \textbf{1. Initialization ($t=0$).} 
At the onset of training, the server initializes $K$ global models $\{M_1, \dots, M_K\}$ with random weights $w_j^{(0)}$, $j \in \{1, \dots, K\}$, representing the starting centroids for the $K$ potential clusters, which correspond to the different device types.

\noindent \textbf{2. Model Broadcast.} 
At the start of every communication round $t$, the server broadcasts the current set of $K$ models to all $N$ participating clients. For the first round, the model weights are initialized randomly; for all subsequent rounds ($t>0$), these weights are the aggregated ones from the previous round, carrying the learned knowledge of their respective clusters.

\noindent \textbf{3. Loss Vector Computation.}
Clients perform a single forward pass on all $K$ received models to compute a loss vector $v_i \in \mathbb{R}^K$. Adhering to our constraints, client $i$ computes the $j$-th component using strictly the reconstruction term on benign data:
\begin{equation}
\mathcal{L}^{i,j}_{\mathrm{MSE}} = \mathcal{L}_{\mathrm{MSE}}\left(\mathcal{D}_i^{\mathrm{benign}}, M_j(x)\right)
\end{equation}
As mentioned previously, the classifier head is frozen during this step to strictly isolate the feature extraction functionality.
Each client sends its computed loss vector to the server.

\noindent \textbf{4. Clustering \& Matching.} 
The server groups the collected vectors into $K$ clusters using K-Means
and performs a minimum-cost bipartite matching between the new clusters and the existing models
(as in \cite{bhatia2025clove}).

\noindent \textbf{5. Model Training.} 
Each client $i$ is assigned to the model $M_j$ corresponding to their cluster. They then perform local training of $M_j$ on the entire dataset $\mathcal{D}_i$ for $\tau$ epochs using the full $\text{DM}^2\text{A}$ architecture (i.e. updating both heads via $\mathcal{L}_{\mathrm{total}}$). 

\noindent \textbf{6. Model Aggregation.} The server aggregates updates solely within each cluster:
\begin{equation}
w_{j}^{(t+1)} \leftarrow \sum_{i \in C_j} \frac{n_i}{n_{C_j}} \left( w_{j}^t - \eta \nabla \mathcal{L}_{\mathrm{total}}^{i,j}\left(w_j^{(t)}\right) \right)
\end{equation}
where $C_j$ is the set of clients assigned to model $j$ at round $t$, $n_i$ is the number of data points of client $i$, $n_{C_j}$ is the total number of data points of all clients assigned to model $j$ at round $t$, and $\eta$ is the learning rate.

\noindent \textbf{7. Stabilization Phase.}
The process repeats until cluster assignments stabilize (i.e. client-to-cluster assignment do not change for 3 consecutive rounds). Once stable, the server ceases the loss vector computation and clustering \& matching steps. The system transitions to efficient parallel instances of federated aggregation (FedAvg), where clients simply receive and train their assigned cluster model, thus drastically reducing communication overhead.

\subsection{Deployment and Inference}

Post-training, the specialized cluster models are deployed to the edge devices. A critical feature of our $\text{DM}^2\text{A}$ architecture is its adaptability to the annotation capabilities of the target environment. The inference logic branches based on whether the specific edge client operates in a labeled or unlabeled setting:

\subsubsection{Scenario I: Labeled Environments} 
In deployment scenarios where the client possesses annotated data or requires specific categorization of threats, the system utilizes the \textbf{Classifier Head} ($C_\theta$). 

The input $x$ is mapped to the latent space $z = E_\phi(x)$, and the classifier outputs the specific class prediction:
\begin{equation}
\hat{y} = \text{argmax}\left(C_\theta(z)\right)
\end{equation}
This mode allows the device to leverage the available labels to identify specific attack types (e.g., Mirai, DoS).

\subsubsection{Scenario II: Unlabeled Environments} 
In scenarios where the client lacks annotations (a common constraint in massive IoT fleets), the system relies purely on the \textbf{Reconstruction Head} ($D_\psi$) to perform anomaly detection.

The input is passed from the Shared Encoder to the Reconstruction Head to generate a reconstructed input $\hat{x} = D_\psi\left(E_\phi(x)\right)$. The system calculates the MSE between the input and the output. If this error exceeds a calibrated threshold $\tau_i$, the traffic is flagged as anomalous:
\begin{equation}
\text{Status}(x) = 
\begin{cases} 
\text{Anomalous} & \text{if } \mathcal{L}_{\mathrm{MSE}}(x, \hat{x}) > \tau_i \\
\text{Normal} & \text{otherwise}
\end{cases}
\end{equation}

\noindent \textbf{Threshold Selection ($\tau_i$)}. 
For clients operating in this unlabeled mode, we employ a localized calibration strategy. Adhering to standard practices in reconstruction-based anomaly detection \cite{saez2023clustered, bhatia_unsupervised_2019}, we derive the decision boundary strictly from the error distribution of normal traffic.
Specifically, we set the threshold to the \textbf{empirical maximum} reconstruction error observed on the client's benign validation $\mathcal{D}_{\mathrm{val}}^{\mathrm{benign}}$ samples:
\begin{equation}
\tau_i = \max_{x \in \mathcal{D}_{\mathrm{val}}^{\mathrm{benign}}} \left(\mathcal{L}_{\mathrm{MSE}}(x, \hat{x})\right)
\end{equation}
This establishes a conservative boundary where any live traffic yielding a reconstruction error higher than the worst-case normal sample is flagged as a potential threat.

\section{Experimental Setup}
\label{sec:experimental_setup}

\subsection{Datasets}

Selecting appropriate benchmarks for this study presented a particular challenge, as in order to best showcase the features and performance of CLAD, we needed network traffic datasets that satisfy strict criteria: a sufficient number of distinct device types, a diverse array of attack types, and adequate sample volume for both types to ensure robust training. Many standard intrusion detection datasets fail to satisfy these requirements concurrently, often lacking the granular device separation or the sufficient data density required for realistic FL. Consequently, we utilize two large-scale datasets that are suitable for examining the different aspects of our proposed framework and the baselines, alongside a third dataset included for preliminary evaluation:

\subsubsection*{CIC IoT-DIAD 2024~\cite{rabbani_device_2025}} This dataset captures a realistic testbed of heterogeneous devices, featuring 33 distinct attacks grouped into seven major categories (e.g., DDoS, Recon, Web-based, Mirai).

\subsubsection*{Gotham 2025~\cite{belarbi2025gotham}} Built on the Gotham testbed, the dataset Gotham 2025 covers modern threats like CoAP Amplification and Remote Code Execution. Notably, its network traffic was collected separately for each IoT device.

\subsubsection*{UNSW~\cite{hamza2019detecting}} We include this dataset as a supplementary benchmark for our initial experiments. It is characterized by severe class imbalance ($>95\%$ benign samples for most devices) and an inherently non-Independent and Identically Distributed (non-IID) nature, evidenced by minimal attack label overlap across device types. Due to the limited sample volume for the majority of attacks ($<40$ samples per device type), it does not meet the density criteria for our primary federated learning experiments and is therefore used only for the initial set of experiments.

\begin{table}[t]
\caption{Detailed comparison of feature availability across datasets.}
\centering
\label{tab:features_comparison}
\renewcommand{\arraystretch}{1.2}
\setlength{\tabcolsep}{4pt}
\begin{tabular}{l l c c c}
\toprule
\textbf{Scope} & \textbf{Feature Type} & \textbf{CIC} & \textbf{Gotham} & \textbf{UNSW} \\
\midrule

\multirow{2}{*}{\textbf{Packet}} 
 & Temporal (Timestamp, Jitter, IAT) & \checkmark & \checkmark & -- \\
 & Headers (TTL, Flags, Window) & \checkmark & \checkmark & -- \\
\midrule

\multirow{3}{*}{\textbf{Flow}} 
 & Volumetric Stats (Means, Vars) & \checkmark & -- & -- \\
 & Aggregated Counts (Pkts, Bytes) & \checkmark & \checkmark & \checkmark \\
 & App. Layer Content (DNS, TLS) & \checkmark & -- & -- \\
\midrule

\multirow{1}{*}{\textbf{Context}} 
 & Directionality (Local/Internet) & -- & -- & \checkmark \\

\bottomrule
\end{tabular}
\end{table}









\subsection{Preprocessing}


\noindent \textbf{Feature Engineering \& Normalization}.
First, we removed leakage identifiers (e.g., IP and MAC addresses) from all datasets. Note that our method does not require device identity information. For CIC IoT-DIAD, we filtered certain columns with ambiguous names, values or labels, retaining 110 features and 6 ``umbrella'' attack categories. For Gotham, we adopt the authors' official workflow to standardize protocols and group fine-grained labels, resulting in 68 features and 5 attack categories. For UNSW, we retained 138 features and utilized the data ``as-is'' without further feature selection. We applied Min-Max scaling to all numeric features on a per-device basis for all datasets. An overview of the dataset features can be seen in Table \ref{tab:features_comparison}.

\noindent \textbf{Filtering \& Splitting}.
To ensure experimental validity, we filtered for devices containing samples for all target attack labels with sufficient volume ($>500$ samples/label). For CIC IoT-DIAD, this criterion yielded 10 devices (Amazon Echo Dot, Amazon Echo Show, Amazon Echo Studio, Amazon Smart Plug, Amcrest WiFi Camera, Arlo Base Station, Arlo Q Camera, Atomi Coffee Maker, Cocoon Smart Fan, Eufy HomeBase 2) covering 6 attacks (DoS, Mirai, Recon, Spoofing, Web-Based, Brute-Force). Similarly, for Gotham, this resulted in 5 qualifying devices (domotic-monitor, building-monitor, combined-cycle, camera-street, camera-museum) covering 5 attacks (Network Scanning, Brute-Force, Infection, C\&C Communication, DoS). For UNSW, we selected the 6 devices (out of 10) that contained at least 3 distinct attack types (TcpSynDevice, Ssdp, TcpSynReflection) to ensure sufficient diversity for validation (TP-Link smart plug, Samsung smartcam, Philips Hue bulb, WEMO Power Switch, WEMO Motion Sensor, Chromecast Ultra). Unless otherwise stated, we enforce a 50:50 Benign/Attack class balance, randomly sample $1000$ records per client, assume IID and fully labeled data and apply a 50:50 train/test split.

\subsection{Data Partitioning}

To rigorously evaluate CLAD, we employ a controlled partitioning strategy designed to disentangle the complex sources of heterogeneity found in IoT networks. While the majority of existing works in federated anomaly detection utilize datasets in their raw, ``as-is" state, such approaches mix different sources of variability, such as label distribution, data quantity, and feature skew, making it difficult to attribute performance gains to specific algorithmic features. In contrast, our approach aims to establish a stable and clean baseline that allows us to examine the specific impact of each experimental variable while holding other conditions constant.

To achieve this, we strictly utilize the subset of ``ground-truth'' source devices identified in the preprocessing phase (10 for CIC IoT-DIAD, 5 for Gotham) that share an identical set of attack labels. From each source device, we generate $5$ non-overlapping client datasets via random sampling, unless otherwise stated. By deriving all $5$ clients for a single cluster from one physical source device, we guarantee intra-cluster homogeneity, ensuring that clients within a cluster are sampled from the exact same underlying distribution. Thus, we can systematically vary parameters such as the number of clients ($N$), local class imbalance, or data volume, without the uncontrolled noise inherent in raw dataset partitioning. As mentioned above, this partitioning does not apply to UNSW, which is used `as-is', given its inherent label imbalance and non-IID characteristics.


\subsection{Baselines}

To the best of our knowledge, there exists no framework that jointly performs anomaly detection (unsupervised) and attack classification (supervised) in either a federated or clustered federated learning setup. Given this absence of direct competitors, we benchmark CLAD against four established strategies that represent the distinct paradigms relevant to this domain, ranging from non-collaborative approaches to federated clustering methods:

\begin{itemize}
    
    \item \textbf{Local:} This approach represents the extreme of fully personalized learning. Each client trains a model exclusively on its local data with no external communication.

    \item \textbf{FedAvg:} We utilize the standard Federated Averaging algorithm~\cite{mcmahan2017communication}, where a single global model is aggregated from all participating clients.
    
    \item \textbf{IFCA:} The Iterative Federated Clustering Algorithm~\cite{ghosh2020efficient} is a prominent clustered FL approach that identifies client clusters by 
    assigning each client to the model (and the corresponding cluster) for which it has the lowest loss across all models.
    To ensure a fair comparison with our framework, we declare stability when clusters do not change for 3 consecutive rounds.

    \item \textbf{CFL-AD:} A CFL baseline for anomaly detection that clusters clients based on the similarity of their model weights~\cite{saez2023clustered}. Adhering to the same convergence strictness as with IFCA above, we evaluate two variations:

    \begin{itemize}
        \item \textit{Standard (\textbf{CFL-ADS}):} Uses the original unsupervised architecture, targeting only anomaly detection.
        \item \textit{Enhanced (\textbf{CFL-ADE}):} Replaces the local model with our proposed $\text{DM}^2\text{A}$ architecture. This allows the baseline to perform joint anomaly detection and attack classification.
    \end{itemize}
    

\end{itemize}

\subsection{Metrics}

We evaluate the efficacy of the proposed framework using a suite of both classification and anomaly detection metrics designed to capture robustness under class imbalance, an inherent characteristic of intrusion detection datasets where benign traffic typically dominates.

\begin{itemize}
    
    \item \textbf{Classification Macro F1-Score (CLS F1):} We report the Macro F1-Score for the multi-class classification task. It is calculated independently for each class and then averaged, ensuring that minority attack classes contribute equally to the score.
    
    \item \textbf{Classification Accuracy (CLS ACC):} This measures the classification accuracy across all classes, encompassing the benign class and the specific attack classes corresponding to each dataset.
    
    \item \textbf{Anomaly Detection F1-Score (AD F1):} We report the F1-Score specifically for the binary anomaly detection task. This metric evaluates the model's ability to make the distinction between any type of attack traffic versus benign traffic.
    
    \item \textbf{Matthews Correlation Coefficient (MCC):} We include the MCC metric for the classification task to provide a robust and balanced measure of quality, particularly given the disparity in class sizes.
    
\end{itemize}

\subsection{Implementation Details}
\label{sec:implementation_details}

\noindent \textbf{Model Architecture.}
To ensure a rigorous evaluation, we fix the neural network backbone across all baselines and our proposed framework, unless otherwise stated. We employ the lightweight $\text{DM}^2\text{A}$ described in Section~\ref{sec:system_model}. The Shared Encoder ($E_{\phi}$) consists of an input layer matching the feature dimension (110 for CIC, 68 for Gotham, 138 for UNSW), followed by three fully connected hidden layers. For the CIC and UNSW datasets, these layers contain 96, 48 and 24 neurons, respectively; for Gotham, they contain 64, 32 and 16 neurons. The Reconstruction Head ($D_\psi$) mirrors the encoder structure. The Classification Head ($C_\theta$) branches from the shared latent representation and comprises two fully connected layers. The first layer reduces the latent dimensionality by a factor of two (24 to 12 for CIC and UNSW, 16 to 8 for Gotham). The final output layer maps this intermediate representation to the target label space.
We utilize GELU activation functions and apply Dropout ($p=0.2$) after each hidden layer. We explicitly opt for these compact configurations to ensure deployability on resource-constrained devices; however, with sufficient computational capability, the model capacity can be increased to further improve performance.

\noindent \textbf{Hyperparameters.}
To ensure fairness, we standardized common hyperparameters across all baselines via grid search. We utilized the AdamW optimizer with  learning rate $\eta=0.01$ and a weight decay of $1e-4$. For the composite loss function, the balancing term was set to $\alpha=0.8$ based on a grid search over the interval [0,1]. The local batch size was set to $32$, and clients performed $\tau=5$ local epochs per communication round. For the federated training process, we ran experiments for a maximum of $R=100$ rounds. We utilized a full client participation rate to eliminate noise from random sampling and isolate convergence behaviors.

We employ the standard model weight averaging for the federated aggregation stage. Evaluation metrics are calculated on each client's test set after each round and averaged across all clients. All reported results are averaged over three random seeds.

\noindent \textbf{Experimental Environment.} 
We implemented our framework and all baselines using the PyTorch library~\cite{paszke2019pytorch}.
All experiments were conducted on a server equipped with a 32-core AMD Ryzen Threadripper PRO CPU, 504 GB of memory and 1.5 TB of storage.

\section{Evaluation Results}
\label{sec:results}

\begin{table}[t]
\centering
\caption{Performance in balanced \& IID settings.}
\label{tab:dataset_baseline_metrics}
\renewcommand{\tabcolsep}{2pt}
\footnotesize
\begin{tabular}{llcccc}
\toprule
\textbf{Dataset} & \textbf{Baseline} 
& \textbf{CLS F1(↑)} & \textbf{CLS ACC(↑)} 
& \textbf{AD F1(↑)} & \textbf{CLS MCC(↑)} \\
\midrule

\multirow{6}{*}{CIC} 
& Local  
  & 0.826 {\scriptsize$\pm$ 0.00}
  & 0.873 {\scriptsize$\pm$ 0.00}
  & 0.952 {\scriptsize$\pm$ 0.04}
  & 0.828 {\scriptsize$\pm$ 0.01} \\
& FedAvg   
  & 0.700 {\scriptsize$\pm$ 0.00}
  & 0.783 {\scriptsize$\pm$ 0.02}
  & 0.868 {\scriptsize$\pm$ 0.03}
  & 0.707 {\scriptsize$\pm$ 0.02} \\
& IFCA        
  & 0.738 {\scriptsize$\pm$ 0.05}
  & 0.806 {\scriptsize$\pm$ 0.03}
  & 0.891 {\scriptsize$\pm$ 0.03}
  & 0.743 {\scriptsize$\pm$ 0.05} \\
& CFL-ADS        
  & --- 
  & --- 
  & 0.749 {\scriptsize$\pm$ 0.02}
  & --- 
  \\
& CFL-ADE        
  & 0.857 {\scriptsize$\pm$ 0.01}
  & 0.901 {\scriptsize$\pm$ 0.01}
  & 0.956 {\scriptsize$\pm$ 0.02}
  & 0.868 {\scriptsize$\pm$ 0.00} \\
& \textbf{Ours}       
  & \textbf{0.875 {\scriptsize$\pm$ 0.00}}
  & \textbf{0.911 {\scriptsize$\pm$ 0.00}}
  & \textbf{0.970 {\scriptsize$\pm$ 0.01}}
  & \textbf{0.880 {\scriptsize$\pm$ 0.01}} \\
\midrule

\multirow{6}{*}{Gotham} 
& Local  
  & 0.888 {\scriptsize$\pm$ 0.02}
  & 0.953 {\scriptsize$\pm$ 0.00}
  & 0.997 {\scriptsize$\pm$ 0.00}
  & 0.936 {\scriptsize$\pm$ 0.00} \\
& FedAvg   
  & 0.856 {\scriptsize$\pm$ 0.07}
  & 0.933 {\scriptsize$\pm$ 0.02}
  & 0.996 {\scriptsize$\pm$ 0.01}
  & 0.912 {\scriptsize$\pm$ 0.02} \\
& IFCA        
  & 0.854 {\scriptsize$\pm$ 0.03}
  & 0.931 {\scriptsize$\pm$ 0.01}
  & 0.997 {\scriptsize$\pm$ 0.00}
  & 0.909 {\scriptsize$\pm$ 0.02} \\
& CFL-ADS        
  & --- 
  & --- 
  & 0.989 {\scriptsize$\pm$ 0.02}
  & --- 
  \\
& CFL-ADE        
  & 0.904 {\scriptsize$\pm$ 0.01}
  & 0.957 {\scriptsize$\pm$ 0.00}
  & 0.998 {\scriptsize$\pm$ 0.01}
  & 0.941 {\scriptsize$\pm$ 0.01} \\
& \textbf{Ours}      
  & \textbf{0.910 {\scriptsize$\pm$ 0.01}}
  & \textbf{0.962 {\scriptsize$\pm$ 0.00}}
  & \textbf{0.999 {\scriptsize$\pm$ 0.00}}
  & \textbf{0.948 {\scriptsize$\pm$ 0.01}} \\
\midrule

\multirow{6}{*}{UNSW} 
& Local  
  & 0.817 {\scriptsize$\pm$ 0.00}
  & 0.850 {\scriptsize$\pm$ 0.01}
  & 0.951 {\scriptsize$\pm$ 0.01}
  & 0.781 {\scriptsize$\pm$ 0.00} \\
& FedAvg   
  & 0.673 {\scriptsize$\pm$ 0.03}
  & 0.801 {\scriptsize$\pm$ 0.00}
  & 0.929 {\scriptsize$\pm$ 0.02}
  & 0.724 {\scriptsize$\pm$ 0.02} \\
& IFCA        
  & 0.751 {\scriptsize$\pm$ 0.00}
  & 0.816 {\scriptsize$\pm$ 0.01}
  & 0.936 {\scriptsize$\pm$ 0.01}
  & 0.735 {\scriptsize$\pm$ 0.02} \\
& CFL-ADS        
  & --- 
  & --- 
  & 0.841 {\scriptsize$\pm$ 0.03}
  & --- 
  \\
& CFL-ADE        
  & 0.811 {\scriptsize$\pm$ 0.02}
  & 0.840 {\scriptsize$\pm$ 0.00}
  & 0.946 {\scriptsize$\pm$ 0.02}
  & 0.785 {\scriptsize$\pm$ 0.01} \\
& \textbf{Ours}      
  & \textbf{0.842 {\scriptsize$\pm$ 0.01}}
  & \textbf{0.870 {\scriptsize$\pm$ 0.00}}
  & \textbf{0.963 {\scriptsize$\pm$ 0.01}}
  & \textbf{0.798 {\scriptsize$\pm$ 0.01}} \\
\bottomrule
\end{tabular}
\end{table}

In this section, we present a comprehensive analysis of CLAD's performance.

\subsection{Balanced \& IID Scenario}

\begin{figure}[t]
    \centering
    \includegraphics[width=\linewidth]{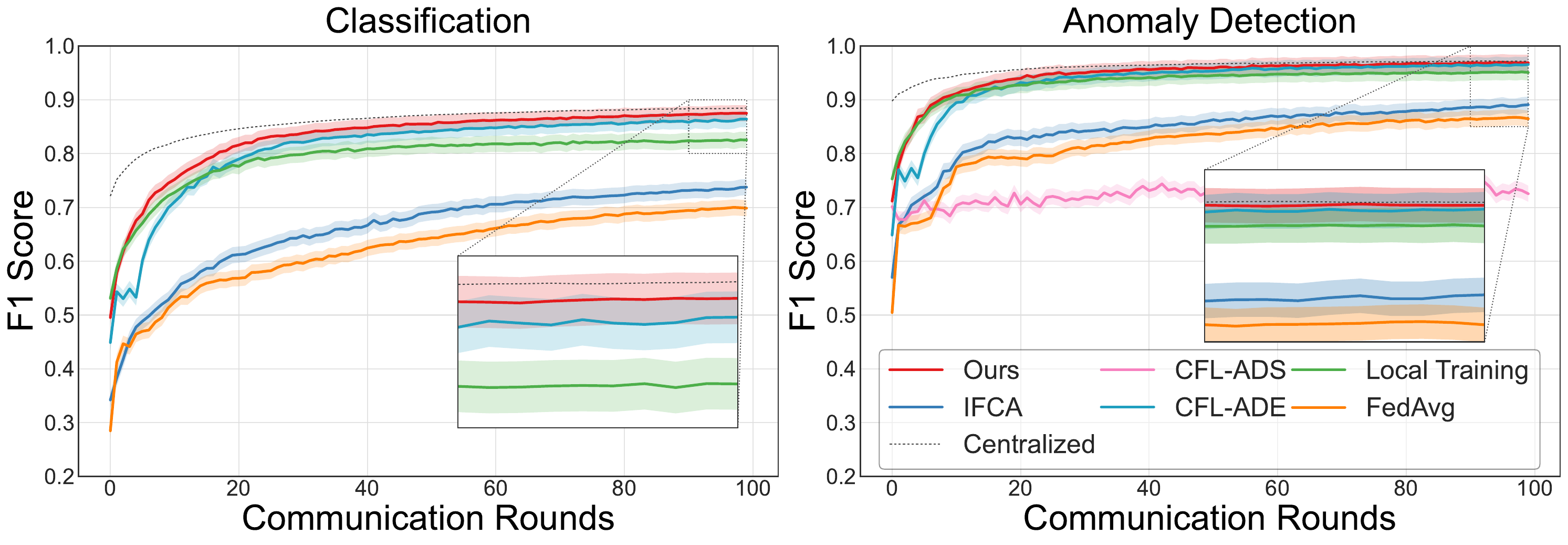}
    \caption{Performance under balanced \& IID scenario for CIC}
    \label{fig:balanced}
\end{figure}

We begin with an ideal scenario designed as a highly controlled environment. In this setup, every client is assigned exactly 1000 samples with a 50:50 Benign/Attack class balance, and all class labels are available to every client. The data is distributed in an IID manner, and each ground-truth cluster (device type) consists of exactly 5 clients. For reference, we also sketch the \textit{Centralized} baseline where all local datasets are uploaded and trained on the server, i.e. with a method that disregards privacy constraints and corresponds to traditional anomaly detection methods. As shown in Fig.~\ref{fig:balanced}, our approach demonstrates rapid convergence, stabilizing near the centralized upper bound (dotted line) within fewer communication rounds compared to other baselines.

Table~\ref{tab:dataset_baseline_metrics} details the performance across the CIC, Gotham and UNSW datasets. Our method achieves the highest metrics across all settings, outperforming the closest clustered baseline enhanced with our model (CFL-ADE). Crucially, the results highlight the severity of negative transfer in heterogeneous environments: both standard FedAvg and IFCA frequently underperform Local (e.g., 0.700 vs 0.826 F1 on CIC), as averaging model weights from functionally distinct devices dilutes model specificity. By effectively clustering these distinct signatures, our framework eliminates this interference, surpassing local training performance by leveraging shared knowledge within valid peer groups.

\begin{table*}[t]
\centering
\caption{Performance under varying label imbalance ratios (percentage of benign samples) and Dirichlet non-IID severity ($\beta$). Smaller $\beta$ indicates stronger heterogeneity.}
\label{tab:merged_imbalanced_non_iid}
\renewcommand{\tabcolsep}{1.5pt}
\footnotesize

\begin{tabular}{l l 
    c c c c c c c c | 
    c c c c c c c c c}
\toprule
\multirow{3}{*}{\textbf{Dataset}} 
& \multirow{3}{*}{\textbf{Baseline}}
& \multicolumn{8}{c|}{\textbf{Label imbalance (benign \%)}} 
& \multicolumn{8}{c}{\textbf{Dirichlet non-IID ($\beta$)}} \\
\cmidrule(lr){3-10} \cmidrule(lr){11-18}
& & \multicolumn{2}{c}{20\%} & \multicolumn{2}{c}{50\%} & \multicolumn{2}{c}{80\%} & \multicolumn{2}{c|}{95\%}
& \multicolumn{2}{c}{$\beta=0.10$} & \multicolumn{2}{c}{$\beta=0.25$} & \multicolumn{2}{c}{$\beta=0.50$} & \multicolumn{2}{c}{$\beta=1.00$} \\
& & CLS F1 & AD F1 & CLS F1 & AD F1 & CLS F1 & AD F1 & CLS F1 & AD F1 
& CLS F1 & AD F1 & CLS F1 & AD F1 & CLS F1 & AD F1 & CLS F1 & AD F1 \\
\midrule

\multirow{6}{*}{CIC}
& Local   & 0.841 & 0.973 & 0.825 & 0.952 & 0.771 & 0.896 & 0.630 & 0.784 
              & 0.436 & 0.856 & 0.670 & 0.923 & 0.715 & 0.931 & 0.784 & 0.947 \\
& FedAvg       & 0.706 & 0.919 & 0.700 & 0.867 & 0.631 & 0.794 & 0.540 & 0.678 
              & 0.506 & 0.786 & 0.671 & 0.841 & 0.664 & 0.847 & 0.686 & 0.859 \\
& IFCA         & 0.749 & 0.933 & 0.737 & 0.891 & 0.696 & 0.823 & 0.560 & 0.696 
              & 0.595 & 0.850 & 0.746 & 0.885 & 0.754 & 0.904 & 0.770 & 0.911 \\
& CFL-ADS         & --- & 0.835 & --- & 0.749 & --- & 0.701 & --- & 0.544 
              & --- & 0.753 & --- & 0.754 & --- & 0.755 & --- & 0.750 \\
& CFL-ADE         & 0.879 & 0.984 & 0.857 & 0.956 & 0.816 & 0.906 & 0.643 & 0.769 
              & \textbf{0.776} & \textbf{0.901} & \textbf{0.831} & \textbf{0.945} & 0.804 & 0.929 & 0.857 & 0.951 \\
& \textbf{Ours} & \textbf{0.888} & \textbf{0.985} & \textbf{0.875} & \textbf{0.970} & \textbf{0.845} & \textbf{0.930} & \textbf{0.714} & \textbf{0.828} 
              & 0.720 & 0.882 & 0.810 & 0.932 & \textbf{0.825} & \textbf{0.938} & \textbf{0.864} & \textbf{0.966} \\
\midrule

\multirow{6}{*}{Gotham}
& Local   & 0.907 & \textbf{0.999} & 0.898 & 0.997 & 0.877 & 0.998 & 0.781 & 0.995 
              & 0.531 & 0.971 & 0.689 & 0.990 & 0.820 & 0.998 & 0.860 & 0.997 \\
& FedAvg       & 0.865 & 0.997 & 0.862 & 0.997 & 0.845 & 0.998 & 0.802 & 0.998 
              & 0.765 & 0.998 & 0.830 & 0.998 & 0.856 & 0.998 & 0.861 & 0.997 \\
& IFCA         & 0.873 & 0.997 & 0.854 & \textbf{0.999} & 0.852 & 0.998 & 0.811 & 0.998 
              & 0.748 & 0.998 & 0.831 & 0.998 & 0.846 & 0.998 & 0.849 & 0.998 \\
& CFL-ADS       & --- & 0.993 & --- & 0.989 & --- & 0.834 & --- & 0.920 
              & --- & 0.985 & --- & 0.989 & --- & 0.990 & --- & 0.989 \\
& CFL-ADE         & 0.905 & \textbf{0.999} & 0.904 & 0.998 & 0.902 & 0.998 & 0.852 & 0.998 
              & \textbf{0.792} & \textbf{0.999} & 0.820 & \textbf{0.999} & 0.788 & 0.998 & 0.857 & 0.998 \\
& \textbf{Ours} & \textbf{0.915} & \textbf{0.999} & \textbf{0.916} & \textbf{0.999} & \textbf{0.913} & \textbf{0.999} & \textbf{0.867} & \textbf{0.999} 
              & 0.732 & 0.991 & \textbf{0.835} & \textbf{0.999} & \textbf{0.864} & \textbf{0.999} & \textbf{0.901} & \textbf{0.999} \\
\bottomrule
\end{tabular}
\end{table*}

\subsection{Label Imbalance}

Real-world IoT traffic is inherently skewed, often dominated by benign activity with only rare attack signatures. We simulate this by varying the percentage of benign samples from 20\% to 95\%, as detailed in the left half of Table~\ref{tab:merged_imbalanced_non_iid}. While extreme sparsity of attack labels generally degrades model sensitivity across all baselines, our framework demonstrates robustness and mitigates the overfitting to the majority class that typically impacts standard supervised approaches.

This robustness is most evident in the extreme 95\% benign scenario. On the CIC dataset, while FedAvg sees its Classification F1 plummet to 0.540 due to the dilution of minority attack gradients, our approach maintains a significantly higher score of 0.714. While CFL-ADE shows consistent performance in less-extreme cases, in the 95\% scenario it shows substantial performance degradation. Similarly, on Gotham, our method sustains near-perfect Anomaly Detection performance (0.999 F1) and leads classification (0.867 F1), outperforming Local training (0.781 F1). These results confirm that our clustered aggregation preserves critical minority class signals that are otherwise lost in global averaging or isolated local training.

\subsection{Non-IID Scenario}

While the previous experiment addressed global class imbalance, this section evaluates performance under statistical heterogeneity \textit{across} clients (non-IID). To model this label skew, we partition the dataset using a symmetric Dirichlet distribution with concentration parameter $\beta \in \{0.1, 0.25, 0.5, 1.0\}$. A lower $\beta$ (e.g., $0.1$) induces extreme heterogeneity where clients may hold samples from only a single class, whereas increasing $\beta$ towards $1.0$ leads to a more uniform partition.

Table \ref{tab:merged_imbalanced_non_iid} (right) reveals that under the most severe heterogeneity ($\beta=0.1$), our method remains highly competitive, ranking a close second to the CFL baseline enhanced by our model (CFL-ADE). This marginal gap can be attributed to the differences in the clustering strategies employed by the two frameworks (CLoVE and CFL-AD). On the CIC dataset, while standard Local training collapses to 0.436 F1, our approach maintains a robust 0.720 F1. Although the extreme fragmentation at $\beta=0.1$ slightly limits the efficacy of our shared reconstruction objective compared to the top baseline, our method still significantly outperforms all remaining baselines, proving its resilience against client drift.

Crucially, our framework demonstrates the fastest recovery rate as heterogeneity decreases. As soon as minimal class overlap emerges at $\beta=0.25$, our method overtakes all competitors. In the Gotham dataset, we outperform all baselines at $\beta=0.25$ with an F1 of 0.835, and extend this lead as the distribution stabilizes, reaching a dominant 0.901 F1 at $\beta=1.0$. This trajectory confirms that while our method is robust in worst-case scenarios, it is uniquely capable of leveraging increasing data quality to maximize performance, unlike baselines that plateau earlier.

\begin{figure}[t]
    \centering
    \includegraphics[width=\linewidth]{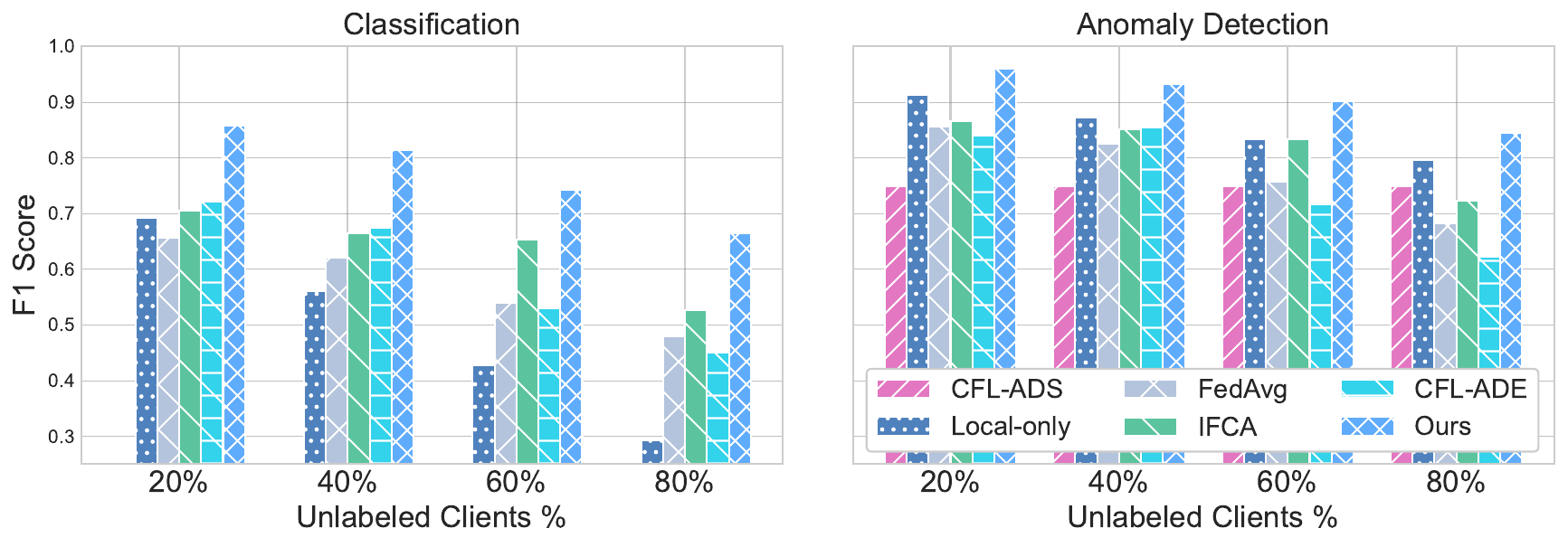}
    \caption{Impact of Unlabeled Client Ratios for CIC}
    \label{fig:unlabeled_mix}
\end{figure}

\begin{table*}[ht]
\centering
\caption{Performance as a function of the number of samples per client and the total number of clients.}
\label{tab:merged_scaling}
\renewcommand{\tabcolsep}{2pt}
\footnotesize

\begin{tabular}{l l 
    c c c c c c c c | 
    c c c c c c c c}
\toprule
\multirow{3}{*}{\textbf{Dataset}} 
& \multirow{3}{*}{\textbf{Baseline}}
& \multicolumn{8}{c|}{\textbf{Samples per client}} 
& \multicolumn{8}{c}{\textbf{Number of clients $N$ (X/Y = Gotham/CIC)}} \\
\cmidrule(lr){3-10} \cmidrule(lr){11-18}
& & \multicolumn{2}{c}{250} & \multicolumn{2}{c}{500} & \multicolumn{2}{c}{1000} & \multicolumn{2}{c|}{2000}
& \multicolumn{2}{c}{10/20} & \multicolumn{2}{c}{25/50} & \multicolumn{2}{c}{50/100} & \multicolumn{2}{c}{100/200} \\
& & CLS F1 & AD F1 & CLS F1 & AD F1 & CLS F1 & AD F1 & CLS F1 & AD F1
& CLS F1 & AD F1 & CLS F1 & AD F1 & CLS F1 & AD F1 & CLS F1 & AD F1 \\
\midrule

\multirow{6}{*}{CIC}
& Local   & 0.714 & 0.888 & 0.782 & 0.924 & 0.825 & 0.952 & 0.858 & 0.969 
              & 0.833 & 0.948 & 0.825 & 0.952 & 0.817 & 0.953 & 0.803 & 0.956 \\
& FedAvg       & 0.628 & 0.819 & 0.667 & 0.844 & 0.700 & 0.867 & 0.704 & 0.888 
              & 0.679 & 0.846 & 0.700 & 0.867 & 0.685 & 0.874 & 0.668 & 0.871 \\
& IFCA         & 0.684 & 0.846 & 0.714 & 0.869 & 0.737 & 0.891 & 0.747 & 0.907 
              & 0.722 & 0.870 & 0.737 & 0.891 & 0.721 & 0.894 & 0.781 & 0.934 \\
& CFL-ADS         & --- & 0.768 & --- & 0.755 & --- & 0.749 & --- & 0.737 
              & --- & 0.736 & --- & 0.749 & --- & 0.758 & --- & 0.770 \\
& CFL-ADE         & 0.789 & 0.916 & 0.843 & 0.939 & 0.857 & 0.956 & 0.881 & 0.974 
              & 0.852 & 0.958 & 0.857 & 0.956 & 0.846 & 0.956 & 0.859 & 0.966 \\
& \textbf{Ours} & \textbf{0.798} & \textbf{0.922} & \textbf{0.856} & \textbf{0.952} & \textbf{0.875} & \textbf{0.970} & \textbf{0.897} & \textbf{0.981} 
              & \textbf{0.864} & \textbf{0.961} & \textbf{0.875} & \textbf{0.970} & \textbf{0.876} & \textbf{0.974} & \textbf{0.873} & \textbf{0.976} \\
\midrule

\multirow{6}{*}{Gotham}
& Local   & 0.849 & 0.998 & 0.879 & 0.997 & 0.887 & 0.996 & 0.904 & 0.997 
              & 0.907 & 0.998 & 0.887 & 0.996 & 0.872 & 0.996 & 0.870 & 0.998 \\
& FedAvg       & 0.864 & 0.997 & 0.862 & 0.996 & 0.855 & 0.997 & 0.834 & 0.998 
              & 0.863 & 0.998 & 0.855 & 0.997 & 0.839 & 0.996 & 0.822 & \textbf{0.999} \\
& IFCA         & 0.857 & 0.998 & 0.866 & 0.997 & 0.854 & 0.997 & 0.846 & \textbf{0.999} 
              & 0.865 & \textbf{0.999} & 0.854 & 0.999 & 0.848 & 0.999 & 0.852 & 0.998 \\
& CFL-ADS   & --- & 0.967 & --- & 0.981 & --- & 0.989 & --- & 0.995 
              & --- & 0.986 & --- & 0.989 & --- & 0.985 & --- & 0.993 \\
& CFL-ADE   & 0.877 & 0.998 & 0.899 & 0.998 & 0.904 & 0.998 & 0.899 & 0.998 
              & 0.911 & 0.998 & 0.904 & 0.998 & 0.893 & 0.998 & 0.891 & \textbf{0.999} \\
& \textbf{Ours} & \textbf{0.883} & \textbf{0.999} & \textbf{0.908} & \textbf{0.999} & \textbf{0.910} & \textbf{0.999} & \textbf{0.918} & \textbf{0.999} 
              & \textbf{0.920} & \textbf{0.999} & \textbf{0.910} & \textbf{0.999} & \textbf{0.902} & \textbf{0.999} & \textbf{0.892} & \textbf{0.999} \\
\bottomrule
\end{tabular}
\end{table*}

\subsection{Label Availability}

In practical IoT ecosystems, obtaining high-quality annotations is resource-intensive; consequently, networks often consist of a large pool of devices whose data completely lacks labels (\textit{unlabeled clients}, i.e. only benign samples available) alongside a small set of clients with labeled data (\textit{labeled clients}). However, efficient frameworks should be able to utilize any amount of label information provided, without completely disregarding it. To evaluate our framework's utility in this semi-supervised regime, we vary the proportion of unlabeled participants from 20\% to 80\%. This setup tests the system's ability to leverage a minority of supervised peers to guide the learning process for the unsupervised majority.

Figure \ref{fig:unlabeled_mix} illustrates the performance on the CIC dataset as label availability becomes scarcer. The results uncover a critical fragility in the baselines. Most notably, CFL-ADE, which consistently ranked as the strong runner-up in fully labeled scenarios, collapses in this semi-supervised regime. As the unlabeled ratio hits 80\%, its Classification F1 drops to $\approx 0.45$, falling behind even the na\"ive FedAvg. This suggests that its clustering performance heavily depends on label supervision, breaking down when the majority of loss vectors lack classification signals. Similarly, Local training degrades substantially to below 0.30 F1, as isolated devices lack the ground truth necessary to form distinct decision boundaries.

In contrast, our method exhibits robust capacity to bridge this gap. Even when 80\% of the network is unlabeled, we sustain a Classification F1-Score of over 0.65 and an Anomaly Detection F1-Score exceeding 0.80, consistently outperforming all baselines. This versatility is intrinsic to our $\text{DM}^2\text{A}$ architecture, which simultaneously optimizes supervised classification and unsupervised reconstruction objectives. This design allows our method to accommodate all clients without making rigid assumptions about label availability: unlabeled clients contribute to feature representation via the unsupervised branch, while ``inheriting" the classification boundaries established by their labeled peers.

\subsection{Scaling with samples}


In real-world IoT deployments, devices often generate sparse traffic patterns due to infrequent user interactions or event-driven behavior~\cite{nguyen2019diot}. Consequently, a robust collaborative system must be capable of converging to a high-performing model even when the local training data is severely limited. Table~\ref{tab:merged_scaling} (left) presents the performance evolution as the number of samples per client increases from 250 to 2000.

The results demonstrate that our approach maintains superior performance in extreme low-data regimes. At the lowest setting of 250 samples, our method achieves a Classification F1 score of 0.798 on CIC and 0.883 on Gotham, significantly outperforming the Local baseline (0.714 and 0.849, respectively). As the sample size scales to 2000, our method consistently retains the top position, reaching an F1 of 0.897 on CIC, confirming its ability to maximize utility from both scarce and abundant data sources.

\subsection{Scaling with clients}

Scalability is a critical requirement for IoT and IIoT networks, where the number of participating clients can vary significantly. Ideally, the learning performance should improve over time regardless of the number of clients. Table~\ref{tab:merged_scaling} (right) analyzes the impact of increasing the fleet size, with client counts ranging from 20 to 200 for CIC and 10 to 100 for Gotham.

The data reveals that our method exhibits remarkable stability compared to local training. As the number of clients increases (implying data is more fragmented across the network), the Local performance tends to degrade; for instance, on the CIC dataset, local Classification F1 drops from 0.833 (20 clients) to 0.803 (200 clients). In contrast, our approach mitigates this fragmentation, maintaining a high F1 score that fluctuates only marginally around 0.873 even at the maximum scale of 200 clients. This consistency confirms that our collaborative mechanism effectively bridges the gap between scattered local datasets, ensuring reliable performance independently of the network size.

\subsection{Communication Efficiency}

Evaluating the trade-off between model performance and communication overhead is critical for deploying federated systems in bandwidth-constrained edge environments. To this end, Table~\ref{tab:efficiency_dominance_compact_3dec} details the performance achieved under challenging data distributions at fixed communication milestones of 13MB and 26MB transmitted per client. Despite all approaches sharing the lightweight model architecture defined in Section~\ref{sec:implementation_details} (33.8K parameters; $\approx 0.129\text{MB}$ at FP32 precision), a stark contrast in efficiency is evident: our method yields better results at the reduced 13MB budget than the best baselines achieve with a full 26MB budget. This is most pronounced in the Label Availability (80\% Unlabeled) scenario, where our method secures a Classification F1 score of 0.609 at 13MB, surpassing the baseline's 0.524 at 26MB. A similar pattern emerges in the Label Imbalance setting for Anomaly Detection, where our 13MB performance (0.784) already exceeds the baselines' maximum capability (0.783).

While our proposed framework delivers a strong average gain of $+5.4\%$ across all tasks, performance varies by scenario. As shown in Table~\ref{tab:efficiency_dominance_compact_3dec}, the non-IID ($\beta=0.25$) setting presents a trade-off, where our method experiences a slight regression compared to the best baseline ($-2.6\%$ in Classification). This indicates that while our approach is highly communication-efficient in unlabeled and unbalanced regimes, extremely heterogeneous distributions may require further tuning. Nevertheless, the aggregate results confirm that our method optimizes the communication budget more effectively than standard approaches, solving complex tasks with a fraction of the data transfer usually required.

\begin{figure}[t]
    \centering
    \begin{subfigure}[b]{\linewidth}
        \centering
        \includegraphics[width=\linewidth]{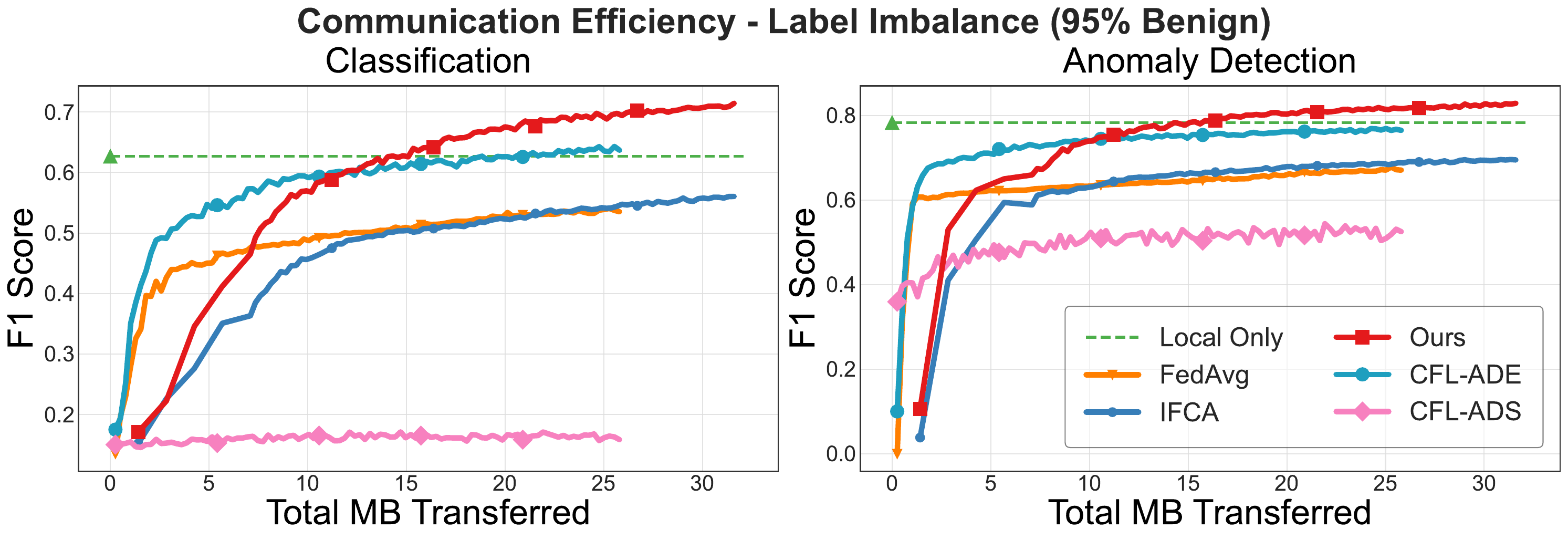}
    \end{subfigure}
    
    
    \begin{subfigure}[b]{\linewidth}
        \centering
        \includegraphics[width=\linewidth]{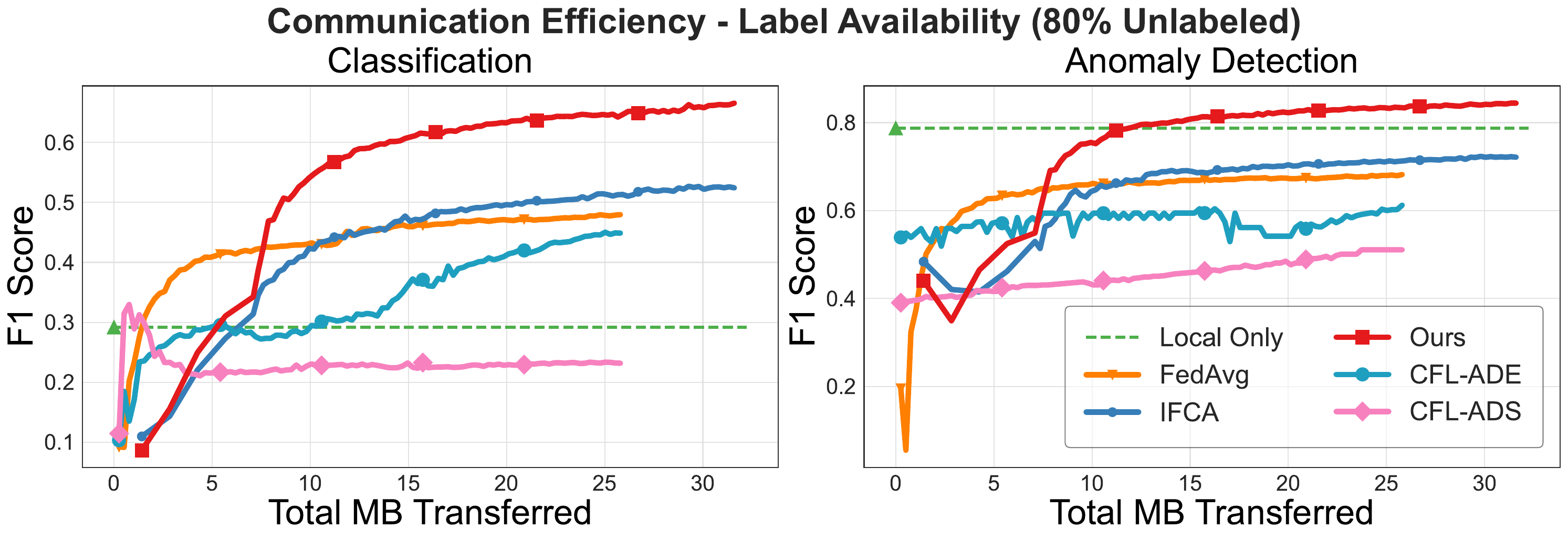}
    \end{subfigure}

    \caption{Communication Efficiency for different scenarios.}
    \label{fig:comm_vs_acc}
\end{figure}

Figure~\ref{fig:comm_vs_acc} illustrates the training dynamics that drive these quantitative gains. In the Label Imbalance and Label Availability scenarios, our method exhibits rapid convergence, establishing a commanding lead within the first few megabytes of transfer. It is important to contextualize these trajectories against the Local benchmark (dashed for reference), which incurs a negligible cost of roughly $0.1$~MB, corresponding to the initial model weights download. While baselines such as FedAvg and IFCA frequently struggle to meaningfully surpass this local performance level, effectively wasting bandwidth, our method justifies the communication overhead by overtaking the local benchmark almost immediately and maintaining a steep upward trajectory throughout the training process.

\begin{table}[t]
    \centering
    \caption{\textbf{Communication Efficiency.}
    Comparison between the best-performing baseline for each scenario and our method under reduced (13MB) and full ($\approx26$MB) communication budgets (per client).
    Improvement denotes the relative gain over the best baseline at the respective budget.}
    \label{tab:efficiency_dominance_compact_3dec}
    \resizebox{\columnwidth}{!}{
    \setlength{\tabcolsep}{3pt} 
    \begin{tabular}{llcccccc}
        \toprule
        \multirow{2}{*}{\textbf{Scenario}} 
        & \multirow{2}{*}{\textbf{Task}} 
        & \multicolumn{2}{c}{\textbf{Best Baseline}} 
        & \multicolumn{2}{c}{\textbf{Ours}} 
        & \multicolumn{2}{c}{\textbf{Relative Gain}} \\
        \cmidrule(lr){3-4}
        \cmidrule(lr){5-6}
        \cmidrule(lr){7-8}
        & & \textbf{@13MB} & \textbf{@26MB} 
          & \textbf{@13MB} & \textbf{@26MB} 
          & \textbf{@13MB} & \textbf{@26MB} \\
        \midrule
        \multirow{2}{*}{\shortstack{Balanced \& IID}}
        & CLS F1 & 0.847 & 0.857 & \textbf{0.855} & \textbf{0.868} & +0.9\% & +1.3\% \\
        & AD F1  & 0.958 & 0.956 & \textbf{0.961} & \textbf{0.966} & +0.3\% & +1.0\% \\
        \midrule
        \multirow{2}{*}{\shortstack{Label Imbalance \\(95\% Benign)}}
        & CLS F1 & 0.608 & 0.637 & \textbf{0.626} & \textbf{0.698} & +3.0\% & +9.6\% \\
        & AD F1  & 0.783 & 0.783 & \textbf{0.784} & \textbf{0.820} & +0.1\% & +4.7\% \\
        \midrule
        \multirow{2}{*}{\shortstack{Non-IID \\ ($\beta=0.25$)}}
        & CLS F1 & \textbf{0.811} & \textbf{0.831} & 0.741 & 0.809 & $-$8.6\% & $-$2.6\% \\
        & AD F1  & \textbf{0.934} & \textbf{0.945} & 0.894 & 0.931 & $-$4.3\% & $-$1.5\% \\
        \midrule
        \multirow{2}{*}{\shortstack{Label Availability \\(80\% Unlabeled)}}
        & CLS F1 & 0.468 & 0.524 & \textbf{0.609} & \textbf{0.656} & +30.1\% & +25.2\% \\
        & AD F1  & 0.788 & 0.788 & \textbf{0.809} & \textbf{0.834} & +2.7\% & +5.8\% \\
        \midrule
        \textit{Average}
        &   &   &   &   &   
        & \textit{+3.0\%} & \textit{+5.4\%} \\
        \bottomrule
    \end{tabular}
    }
\end{table}

\subsection{Computational Impact} 


\begin{figure}[t]
    \centering
    \begin{subfigure}[b]{\linewidth}
        \centering
        \includegraphics[width=\linewidth]{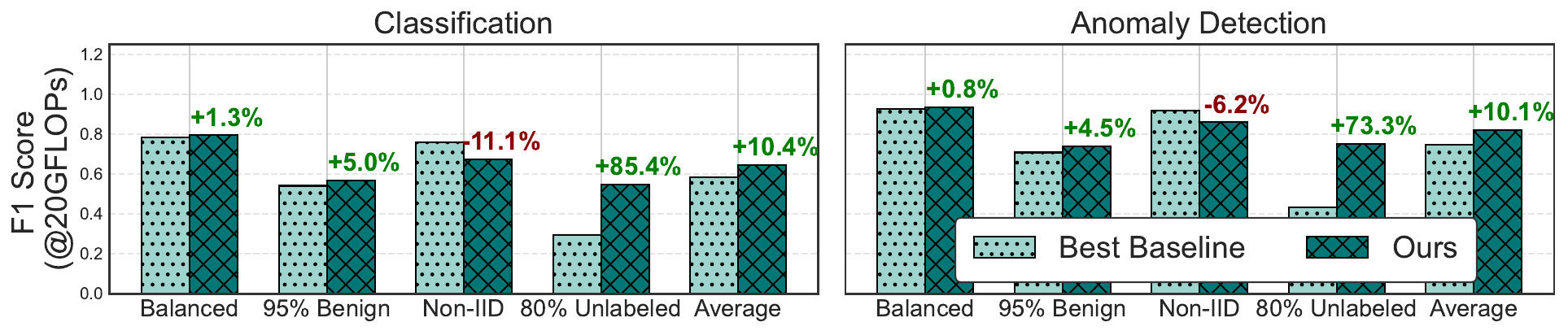}
        \caption{Performance at limited computational budget (20GFLOPs total)}
        \label{fig:20gflops}
    \end{subfigure}

    \centering
    \begin{subfigure}[b]{\linewidth}
        \centering
        \includegraphics[width=\linewidth]{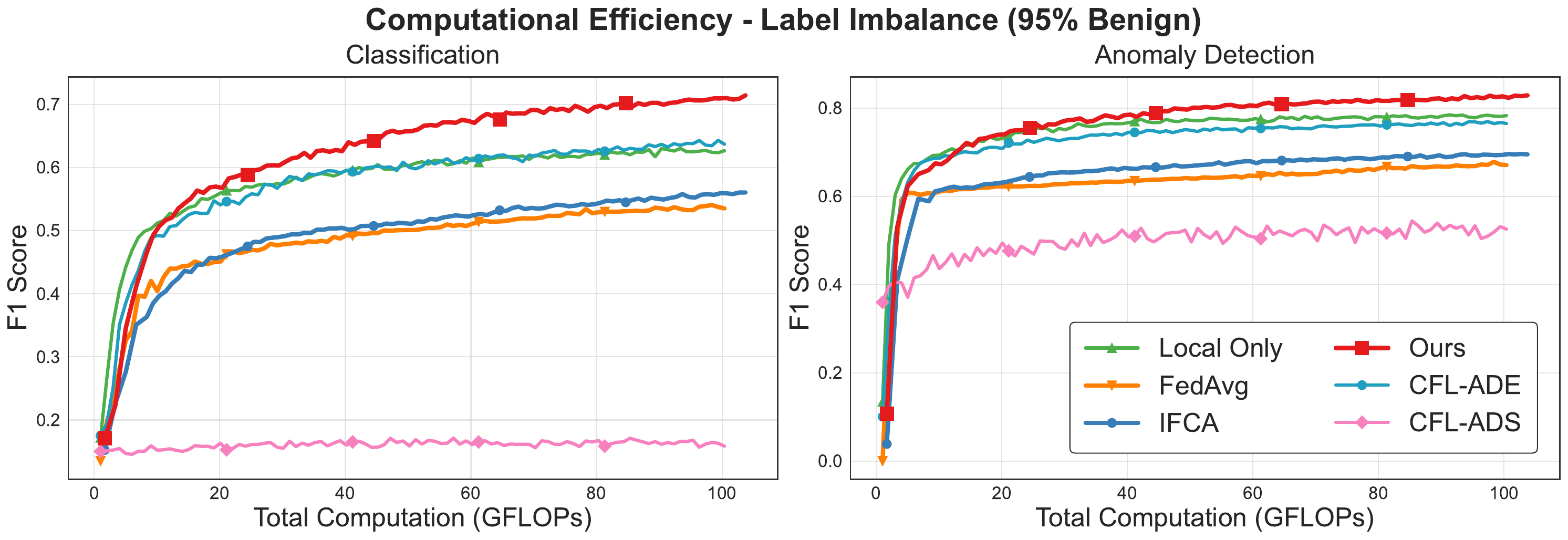}
    \end{subfigure}
    
    
    \begin{subfigure}[b]{\linewidth}
        \centering
        \includegraphics[width=\linewidth]{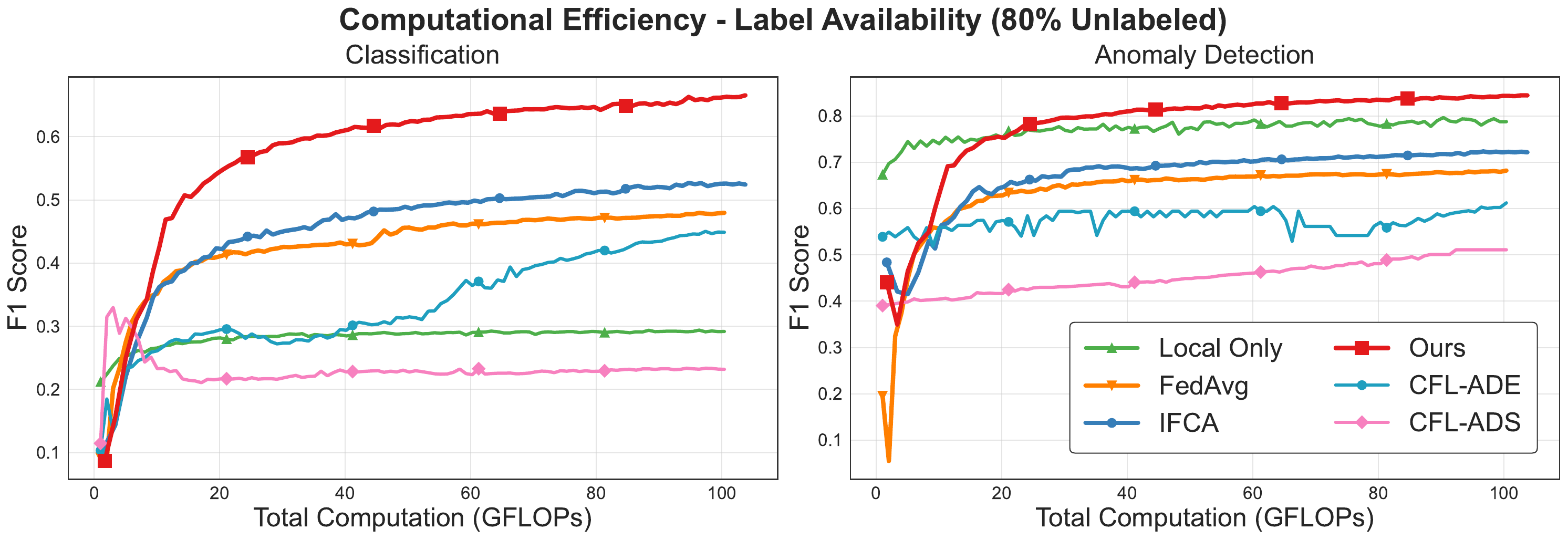}
        \caption{Label Imbalance \& Label Availability Scenarios}
        \label{fig:comp_scenarios}
    \end{subfigure}

    \caption{Computational Impact for different scenarios.}
    \label{fig:comp_vs_acc}
\end{figure}

Iterative clustering algorithms such as the one we employ are frequently viewed as computationally expensive, as they typically impose an additional burden on clients through the extra forward passes required for cluster assignment. Our results demonstrate that this overhead does not hinder training speed in practice, and that our framework utilizes computational resources significantly more effectively than standard approaches. Figure~\ref{fig:20gflops} compares performance at a restricted budget of 20~GFLOPs. Despite the theoretical overhead, our method achieves a substantial average improvement of +10.4\% in Classification and +10.1\% in Anomaly Detection over the best-performing baseline. This gain is most dramatic in the Label Availability (80\% Unlabeled) scenario, where we observe a massive relative improvement of +85.4\% in Classification F1 score, proving that our targeted clustering accelerates feature learning far faster than the cost of the additional passes slows it down.

Figure~\ref{fig:comp_scenarios} further illustrates this ``performance-per-FLOP'' dominance. In the Label Imbalance and Label Availability scenarios, similar to the Communication Efficiency section, our method exhibits a steep vertical trajectory, minimizing the computational waste associated with slow convergence. A striking finding is observed in the unlabeled classification task: our approach achieves superior performance using only 20~GFLOPs than competing baselines achieve after consuming their entire 100~GFLOPs budget. While extremely heterogeneous environments (non-IID) introduce a slight computational trade-off due to the difficulty of clustering divergent data, the overall trend confirms that our method yields a higher return on computational investment, making it uniquely suitable for resource-constrained devices where every FLOP counts.






\section{Conclusion}
\label{sec:conclusion}

In this paper, we introduced CLAD, a novel framework that unifies CFL with $DM^2A$ to overcome device heterogeneity and label scarcity in IoT/IIoT security. Moving beyond the rigid ``one-model-fits-all" paradigm of traditional FL, our label-agnostic framework seamlessly integrates supervised and unsupervised learning to ensure operational resilience; it utilizes every piece of available information so that no label is discarded and no device, regardless of its data availability, is excluded from the collective defense. Extensive evaluations demonstrate that CLAD significantly outperforms state-of-the-art baselines, particularly in label-scarce environments; notably, we achieved a 30\% relative improvement in detection performance with 80\% unlabeled clients, while reducing communication overhead by half. These results confirm that our hybrid architecture offers a scalable, data-efficient solution for robust network intrusion detection across complex IoT ecosystems.


\bibliographystyle{IEEEtran}
\bibliography{references}

\end{document}